\documentclass[letterpaper, 10 pt, conference]{ieeeconf}
\IEEEoverridecommandlockouts

\usepackage{times}
\usepackage{comment}
\usepackage{color}
\usepackage{graphicx}
\usepackage{booktabs}
\usepackage{amsmath}

\DeclareMathOperator*{\argmax}{arg\,max}

\begin{document}
\title{\LARGE \bf Predicting Extubation Readiness in Extreme Preterm Infants based on Patterns of Breathing*}

\author{Charles C. Onu$^{1}$, Lara J. Kanbar$^{2}$, Wissam Shalish$^3$, Karen A. Brown$^{4}$, Guilherme M. Sant'Anna$^3$, \\ Robert E. Kearney$^2$ and Doina Precup$^{1}$% <-this % stops a space
\thanks{*Research supported in part by the Canadian Institutes of Health Research. The work of L. Kanbar was supported in part by the Natural Sciences and Engineering Research Council of Canada. K. Brown was supported in part by the Queen Elizabeth Hospital of Montreal Foundation Chair in Pediatric Anesthesia. }% <-this % stops a space
\thanks{$^{1}$C. C. Onu and D. Precup are with the School of Computer Science, McGill University, Montreal, QC H3A 2B4, Canada. (e-mail: { charles.onu@mail.mcgill.ca; dprecup@cs.mcgill.ca)}}%
\thanks{$^{2}$L. J. Kanbar and R. E. Kearney are with the department of Biomedical Engineering, McGill University, Montreal, QC H3A 2B4, Canada (e-mail: { lara.kanbar@mail.mcgill.ca; robert.kearney@mcgill.ca)}}%
\thanks{$^{3}$W. Shalish and G. M. Sant'Anna are with the division of Neonatology, McGill University Health Center, Montreal, QC H3A 2B4, Canada. }%
\thanks{$^{4}$K. Brown is with the department of Anesthesia, McGill University Health Center, Montreal, QC H3A 2B4, Canada.}%        
}


\maketitle

\begin{abstract}
Extremely preterm infants commonly require intubation and invasive mechanical ventilation after birth. While the duration of mechanical ventilation should be minimized in order to avoid complications, extubation failure is associated with increases in morbidities and mortality. As part of a prospective observational study aimed at developing an accurate predictor of extubation readiness, Markov and semi-Markov chain models were applied to gain insight into the respiratory patterns of these infants, with more robust time-series modeling using semi-Markov models. This model revealed interesting similarities and differences between newborns who succeeded extubation and those who failed. The parameters of the model were further applied to predict extubation readiness via generative (joint likelihood) and discriminative (support vector machine) approaches.  Results showed that up to 84\% of infants who failed extubation could have been accurately identified prior to extubation.

%Extremely preterm infants require respiratory support through intubation and invasive mechanical ventilation after birth. The duration of ventilation should be minimized in order to prevent the development of chronic lung disease. However, premature extubation must be avoided as it could lead to mortality. We applied Markov chain-based modeling to gain insight into the respiratory patterns of preterm newborns. The objective was to develop adequate predictors of extubation readiness. We demonstrated that semi-Markov chain models are more robust than Markov chain models when analysing time-series data. The developed semi-Markov chain model revealed interesting similarities and differences between newborns who succeeded extubation and those who did not. The parameters of the model were further applied in predicting extubation readiness via generative (joint likelihood) and discriminative (support vector machine) approaches. Results show that up to 84\% of infants who failed extubation could have been accurately identified ahead of time.
\end{abstract}

\section{Introduction}

Due to immature lungs and respiratory control, extremely preterm infants (gestational age ≤ 28 weeks) are at high risk of respiratory failure after birth. For that reason, most require endotracheal intubation and Invasive Mechanical Ventilation (MV) within the first hours or days of life \cite{stoll2015trends}. As the respiratory status improves, MV is continuously weaned until the medical team deems that removal of the tube, termed \textit{extubation}, can be performed. This is a difficult decision since if done too early, it could result in the need for re-intubation, a technically challenging intervention that has been associated with increased morbidities and mortality \cite{manley2016extubating,chawla2017markers}. On the other hand, delayed extubation, and hence prolonged mechanical ventilation, could increase the risk of complications. The most common complication is broncho-pulmonary dysplasia (BPD), a form of chronic lung disease associated with serious long-term sequelae \cite{walsh2005extremely}.

%Infants born extremely preterm (gestational age $\leq$ 28 weeks) are usually faced with a high risk of respiratory failure due to immature lungs. Most require EndoTracheal Tube-Invasive Mechanical Ventilation (ETT-IMV) within the first days of life \cite{walsh2005extremely}. This entails a breathing tube inserted into the infant's trachea and connected to a ventilator. The newborn is monitored until an attending clinician deems that {\em extubation} (removal of the tube) should be performed. Time of extubation is a critical decision for clinicians. If done too early, it could result to the need for re-intubation, which is a technically challenging due to inflammation in the trachea, and can further result in morbidity or mortality \cite{baisch2005extubation}. On the other hand, delayed extubation could result to the development of broncho-pulmonary dysplasia (BPD), a severe chronic lung disease \cite{walsh2005extremely}.

This project is part of an ongoing multicenter, multidisciplinary collaborative study aiming to develop an automated prediction tool of extubation readiness using analysis of cardiorespiratory signals in extremely preterm infants (Clinicaltrials.gov identifier: NCT01909947) \cite{shalish2017prediction}. In the present work, we approach the task of predicting whether a patient would succeed or fail extubation based solely on respiratory patterns extracted from a 5-minute period of spontaneous breathing trial (SBT), while patient was still intubated but not receiving mechanical inflations from the ventilator.

%In this work, we approach the task of predicting whether a patient would succeed or fail extubation based on respiratory patterns extracted from a 5-minute period of spontaneous breathing trial (SBT). During SBT, the patient is still intubated, but without active support from the mechanical ventilator. The clinician observes the patient, in order to ensure they are capable of breathing on their own. If so, the clinician proceeds immediately to extubation. If the patient is in distress during this procedure, the ventilator is reconnected. Patients who undergo extubation may need to be reintubated at a later time, due to repeated and prolonged apneas, inadequate blood oxygen levels, or other causes. This study is part of a project whose goal is to provide quantitative input to the physician in order to improve their ability to predict correctly which patients will succeed or fail extubation.

Methods of generative modeling using Markov and semi-Markov chains were applied. The respiratory time series was processed as a sequence of five unique and mutually exclusive breathing patterns/states: Pause, Asynchronous breathing, Movement artifact, Synchronous breathing and Unknown, using an existing method of analysis of the respiratory signals called AUREA (Automated Unsupervised Respiratory Event  Analysis)  \cite{aurearef}. The models fitted to these states sequences revealed key similarities and differences between patients who succeeded and failed extubation. The characteristics were exploited in a support vector machine-based discriminative classifier and results indicated that our model could identify infants who went on failing extubation with up to 84\% accuracy.

%We applied the method of generative modeling using Markov and semi-Markov chains.  For this purpose, the respiratory time series was processed as a sequence of five unique and mutually exclusive breathing patterns/states: Pause, Asynchronous breathing, Movement artifact, Synchronous breathing and Unknown, using an existing method called AUREA \cite{aurearef}. The models fitted to this state sequence revealed key similarities and differences between patients who succeeded and failed extubation. These characteristics were exploited in a support vector machine-based discriminative classifier. Results indicate that our model can identify infants who are not ready for extubation with up to 84\%  accuracy.

The rest of this paper is organized as follows. Section \ref{section_related_work} reviews related work. Section \ref{section_data} describes the data acquisition and preprocessing steps. Section \ref{section_method} describes our methodology. Section \ref{section_results} presents results and provides a discussion on their significance, and in Section \ref{section_discussion}, we discuss future research directions.

\section{Related Work}
\label{section_related_work}
The use of Markov chain models for understanding sequence data and making predictions about outcomes is popular across many domains, especially when the values of the time-series are drawn from a discrete set of states. Ye et al \cite{ye2000markov} applied Markov chain modeling to detect anomalous activities in computer networks. In another study by Gabriel et al \cite{gabriel1962markov}, this method was successfully used to track rainfall occurrence patterns. More recently, Alinovi et al \cite{Alinovi2017} applied semi-Markov chains to model the respiratory rate of infant patients experiencing breathing-related disorders such as apneas. In that study, respiratory rate was transformed from a real-valued time-series in to 5 levels by quantization. The authors demonstrated that semi-Markov chain models accurately described respiratory rate and simulated realistic sequences of respiratory rate in both normal and apneic infants.

Previous work in predicting extubation readiness of extreme preterm newborns have demonstrated that cardiorespiratory signals hold useful discriminating information between infants that succeed and those that fail extubation. In particular 2 prospective studies by Kaczmarek et al \cite{kaczmarek2013heart} and Robles-Rubio et al \cite{roblesrubio2015}on a small cohort of extremely preterm infants, extracted measures of heart rate variability (HRV) and respiratory variability (RV) from electrocardiogram (ECG) and respiratory inductive plethysmography (RIP) signals, respectively. Results revealed that measures of HRV and RV were excellent at detecting extubation successes (i.e. high specificity), but not as sensitive in detecting failures. In \cite{precup2012}, using patients of the same cohort, a Support Vector Machine (SVM) was developed to predict extubation readiness directly from the combination of cardiorespiratory  variability measures. This system achieved improved ability to detect extubation failures (sensitivity of 83\%)  while maintaining a fairly high specificity of 74\%. It should however be noted that these works were carried out on fairly small datasets.

%It was found that HRV and RV were significantly lower in infants that failed extubation, allowing for high sensitivity (100\%) but low specificity (53\%) in detection. In \cite{precup2012}, a Support Vector Machine was developed to predict extubation readiness directly from automatically computed cardiorespiratory variability metrics of a cohort of 53 infants. This system achieved improved specificity of 73.6\% while maintaining a high sensitivity of 83.2\%. It should be noted that these works were carried out on fairly small datasets.

In this current work, we analyzed a larger dataset of 186 babies collected as part of a multi-institutional study. We explore the predictive potential, not of HRV or RV but, of respiratory pattern sequences of the infants. The current work: 1) empirically demonstrated that the semi-Markov modeling has a more robust capability for time-series data than the Markov chain models;
2)	used semi-Markov chain models to understand the transition behavior of meaningful respiratory patterns representing actual respiratory states (not just quantized real-values);
3)	showed that, in addition  to  generative  classification via maximizing joint likelihood, the parameters of semi- Markov chains can be exploited in discriminative classifiers to improve performance. To the best of our knowledge, this is the first work that applies a time-series-based machine learning method to predict extubation readiness in extremely preterm infants.

\section{Data}
\label{section_data}

\subsection{Data Acquisition}
Data from  186 infants  was available for this study. Patients were enrolled and studied in five sites in Canada (Royal Victoria Hospital, Montreal Children's Hospital, Jewish General Hospital) and the USA (Detroit Medical Center, MI, and Women and Infants Hospital of Rhode Island, RI). 
Ethical approval was obtained from each institutional board, and informed parental consent was obtained before recruitment.

\begin{figure}[t]
   \centering
   \includegraphics[width=0.5\textwidth]{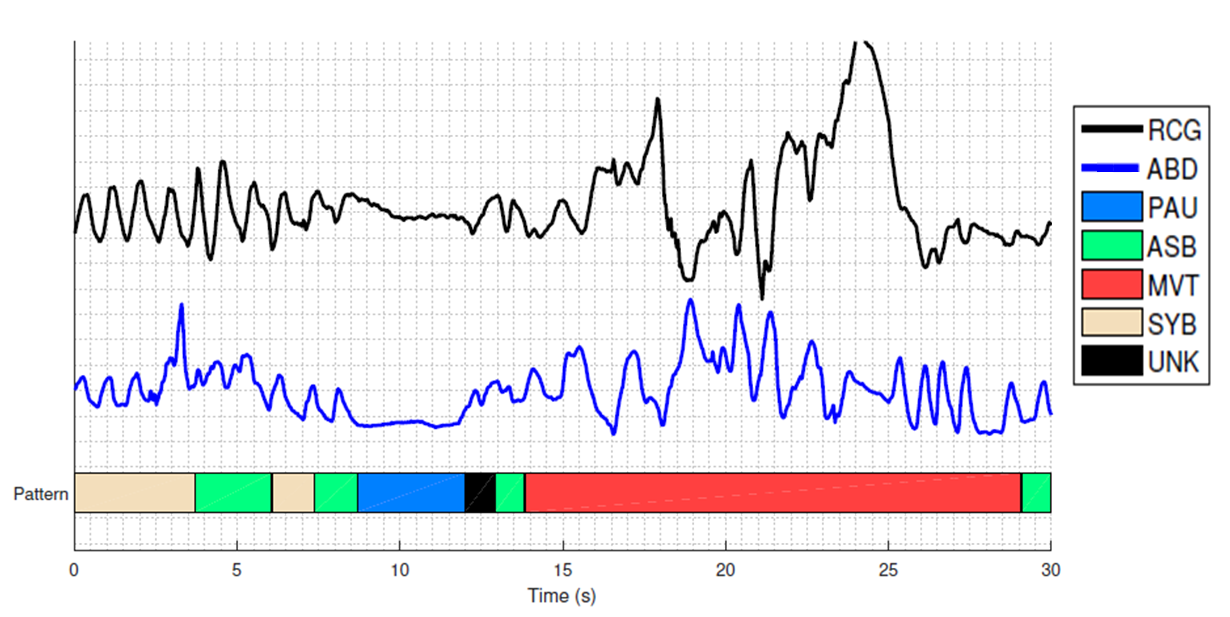}
   \caption{Example of a RCG and ABD signal segment and the corresponding respiratory patterns (PAU, SYB, ASB, MVT, UNK) computed by AUREA.}
   \label{fig_rip_sample}
\end{figure}

Eligible infants were of birth weight $\leq$ 1250g, and receiving MV at time of recruitment. Infants were excluded if they had any major congenital anomalies such as heart disease, or were receiving any vasopressor or sedative drugs at the time of extubation.

Respiratory signals were measured using RIP bands placed around the infant's ribcage and abdomen. Signals were acquired for a 5-minute period of spontaneous breathing  without any mechanical inflation from the ventilator, prior to extubation, at a sampling frequency of 1000Hz. Only infants which were then extubated were included in the dataset. Failure after extubation was defined as re-intubation within 7 days.
More details on the data collection procedure for this study are available in \cite{shalish2017prediction}.

\subsection{Data Preprocessing}

RIP signals sampled at 50Hz were analyzed using AUREA, which extracts sample-by-sample metrics of respiratory power, synchrony between the ribcage (RCG) and abdomen (ABD), and movement artifact \cite{aurearef}. AUREA uses k-means clustering with windowing and smoothing to assign each sample to one of the following respiratory patterns:
\begin{itemize}
\itemsep0em
\item Pause (PAU): A cessation of breathing.
\item Synchronous Breathing (SYB): RCG and ABD are in phase.
\item Asynchronous Breathing (ASB): RCG and ABD are out of phase.
\item Movement Artifact (MVT): Associated with infant moving or nurse handling.
\item Unknown (UNK): Ambiguous patterns not belonging to any other pattern category.
\end{itemize}

AUREA provides repeatable results with no human intervention \cite{aurearef}. An example of RIP signals and corresponding patterns assigned by AUREA to the different samples is shown in Fig. \ref{fig_rip_sample}. 

%anonymised
%.

%Their results showed that as a statistical model, Markov chains provide a "reliable and realistic method to simulate breathing patterns and respiratory pauses/apneas". 

\section{Methods}
\label{section_method}

We applied  Markov chains and Semi-Markov chains to model the time series of respiratory patterns, and then used SVM with features extracted from these chains to predict extubation readiness. We now describe the analysis methods.

\subsection{Discrete-time Markov Chain}

A Markov chain is a probabilistic graphical model in which every node in the chain is only dependent on the one preceding it.  Time-series data can be represented as a Markov chain whereby the nodes of the chain are the values observed at every time step. We use discrete-time finite Markov chains, in which at time $t$, state $x_t$ takes a value from a finite set of states $S$ (in our case, the 5 respiratory patterns provided by AUREA).

%\begin{figure}[b]
%   \centering
%   \includegraphics[width=0.6\textwidth]%{figs/illustration_markov_chain.png}
%   \caption{Example of a Markov chain}
%   \label{fig_illustration_markov_chain}
 %\end{figure}
 
%When used to model time-series data, the implicit assumption of a Markov chain is that the value of an observed variable at a time step, $t$, captures everything there is to know about all preceding time steps from $t-1$ to $1$.

A Markov chain model has 2 sets of parameters: the probability distribution over initial states (a vector) $\pi$ and the transition probabilities between states (a matrix) $A$. Fitting or learning the model of a Markov chain involves estimating these  parameters from data. Their maximum likelihood estimates are given by  \cite{Barbu2008}:
\begin{eqnarray}
\label{eqn_start_dist}
\pi_j &=& \frac{\#\ of\ sequences\ starting\ in\ j}{\#\ of\ sequences}	\quad	\forall j \in S\\
%\label{eqn_transition_prob}
%A_{i,j} &=& \frac{\#\ of\ i\ to\ j\ transitions}{\#\ of\ times\ i\ was\ visited} 	\quad	\forall i,j \in S\\
\label{eqn_transition_prob}
A_{i,j} &=& \frac{n_{ij}}{\sum_j{n_{ij}}} \quad \forall \  i,j \in S 
\end{eqnarray}

where $n_{ij}$ is the number of time steps in which a transition from state i to j occurred.
Given a time-series of observations $x_1, x_2,...,x_T$, the joint likelihood of the sequence according to the Markov chain is given by:
\begin{equation}
\label{eqn_joint_prob_markov}
P(x) = P(x_1) \prod_{t=2}^T P(x_t | x_{t-1}) = \pi_{x_1} \prod_{t=2}^T A_{x_{t-1},x_t}
\end{equation}
Note that in practice, the start state distribution $\pi$ was not included in our model, due to an adapation or transition phase from MV with inflations to ETT-CPAP without inflations.

In order to apply Markov chain models for classification, separate transition models $A^s$, $A^f$ were first fit to the data coming from the success and failure patients, respectively. The classification for a new sequence $x$ is done by computing its posterior likelihood with respect to both models (using Eq.\ref{eqn_joint_prob_markov}), and selecting the class, c whose model gives the higher likelihood:

\begin{equation}
\label{eqn_markov_classification}
\argmax_c L(x |A^c)
\end{equation}

where 
\begin{equation}
\label{eqn_markov_likelihood}
L(x|A^c) = \prod_{t=2}^T A_{x_{t-1},x_t}^c
\end{equation}

%In our problem, we explore the hypothesis that the respiratory patterns of infants who succeed extubation follows a different Markov chain from those who fail. Thus, conditional models are learned by fitting one Markov chain ($\pi_j^s$, $A_{i,j}^s$) to the respiratory pattern sequences of success patients, and another ($\pi_j^f$, $A_{i,j}^f$) to the sequences of failure patients. The prediction for a new example sequence $x$ is made by assigning to it the class (success or failure) of maximum posterior probability. To do this, the likelihood of the sequence is computed with respect to the success and failure models using equation (\ref{eqn_joint_prob_markov}). The class which gives higher likelihood is selected.

\subsection{Discrete-time Semi-Markov Chain}
\label{section_semi_markov_chain}
A Semi-Markov chain model, on the other hand, is characterised by 3 parameters: a start state distribution vector $\pi$; the transition matrix $A$, which stores only cross-state transition probabilities (diagonal elements are 0); and a set of {\em dwell or \textit{sojourn} time distributions} $F$, which model the duration spent in each state until a transition out of that state occurs.

%in that self-transitions, i.e. transitions from a state to itself, are collapsed. Instead, each state has a separate {\em dwell or sojourn time distribution}, which models the duration spent in the state until a transition out of the state occurs, coupled with a probability distribution of transitioning to {\em other} states. The latter results in a chain where every transition results in a state that differs from its predecessor.

This framework was useful in our application for several reasons. First, Markov chains implicitly model dwell times as an exponential distribution \cite{christopher2006pattern} which could introduce bias into the model if underlying data is not actually exponential. Secondly, in data characterised by very long dwell times, the transition probabilities of cross-state transitions (off-diagonal elements) go to 0, making it very difficult to get any useful information from the model. Finally, a Markov chain is highly susceptible to changes in the sampling rate of the data. Semi-Markov chains address all of these issues.

%For example, if the sampling rate were to double, assuming that this does not affect the state labeling, the self-transitions of a Markov chain that works at the sampling rate would roughly double; in contrast, the semi-Markov chain would still have the same dwell time distribution and same probability of transitioning from a state to its successors.

The joint likelihood of a sequence of observations under a semi-Markov chain is given by:
\begin{equation}
\label{eqn_joint_prob_semimarkov}
P(x) = \pi_{x_1} \prod_{t=2}^T A_{x_{t-1},x_t} F_{x_t}(|x_t|)
\end{equation}
where $F_{x_t}(|x_t|)$ is the probability of sojourning in the state $x_t$ for the duration $|x_t|$. ~\cite{puterman1994markov}

In modeling the infant respiratory pattern sequences as a semi-Markov chain, $A$ was learned as before. To fit the dwell time distributions, all dwell times in a state (e.g., PAU) for a particular population (success or failure) were obtained. Several known probability distributions were fit to this data using MATLAB \cite{allfitdist}. The distribution which minimized the Bayesian Information Criterion (BIC) \cite{schwarz1978estimating} was selected. These steps were repeated for all states in both success and failure groups to obtain 10 separate dwell time distributions. Classification of a new example sequence as success or failure was done as described in the previous section by selecting the class of larger posterior likelihood.

%To learn the parameters $\pi$ and $A$, equations \ref{eqn_start_dist} and \ref{eqn_transition_prob} were employed. To model the dwell time distributions, the following process was repeated for the success and then for the failure patients: all dwell times in each state were collected across the patients in the population. A probability density function (PDF) estimate was obtained by plotting a histogram of the data normalised by number of observations in a bin and the bin width. Several known probability distributions were then fit to the data, selecting the one that minimised the Bayesian Information Criterion (BIC) \cite{schwarz1978estimating}. 

%As in the case of Markov chains, prediction for a new sequence $x$ can be done by fitting conditional models (based on success and failure examples) to the data, estimating the likelihood of the new sequence given the models, and selecting the class with the maximum posterior probability for that sequence.

\begin{figure}[b]
  \centering 
  \includegraphics[width=0.49\textwidth]{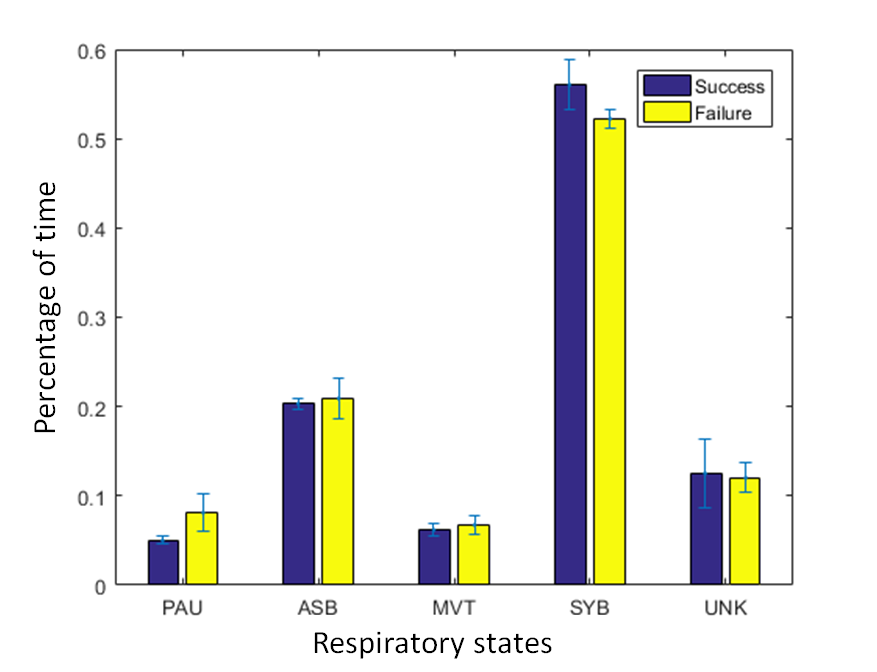} 
  \caption{Percentage of time spent in each respiratory state by success and failure patients during a 5-min period of spontaneous breathing prior to extubation}
  \label{fig_state_duration} 
\end{figure}

\subsection{Support Vector Machine (SVM)}

Using the Markov model likelihood for classification can be sub-optimal if the model structure or some of the model parameters are imprecise. Discriminative models do not make probabilistic assumptions about how the inputs were generated, but rather attempt to learn a (linear or non-linear) boundary between the groups. Support vector machines (SVMs), in particular, learn a maximum margin decision boundary \cite{cortes1995support}. In order to compare our results to the Markov chain case, we derived summary statistics from the respiratory pattern sequence of each patient and used these as inputs to train an SVM. In particular, a radial basis function (RBF) SVM was used. The hyperparameters of the RBF SVM - box constraint, C (which penalises the error function to manage overfitting) and kernel scale $\gamma$ (which controls the width of the Gaussian) - were optimized using leave-one-out cross-validation and the balanced loss metric, as motivated in section \ref{subsec_model_eval}.

\subsection{Symmetric KL Divergence}
The Kullback-Leibler (KL) divergence \cite{kullback1997information} is a measure of how well a distribution $Q$ is approximating another distribution $P$. It is defined as:
\begin{equation}
D_{KL}(P||Q) = \sum_n P_n \log\frac{P_n}{Q_n}
\end{equation}
The KL-divergence is non-symmetric: $D_{KL}(P||Q) \ne D_{KL}(Q||P)$, which is not desirable in our application. Hence, we use symmetrized KL-divergence  to compare distributions over transitions between patterns:
%$$
%D_{KLS}(P||Q) = D_{KL}(P||Q) + D_{KL}(Q||P)
%$$
\begin{equation}
\label{eq_symmetric_kl}
D_{KLS}(P||Q) = D_{KL}(P||Q) + D_{KL}(Q||P)
\end{equation}

\begin{equation}
D_{KLS}(P||Q) = \sum_n (P_n - Q_n) \log\frac{P_n}{Q_n}
\end{equation}

\subsection{Model Evaluation}
\label{subsec_model_eval}
Due to the small size of our dataset and class imbalance, it was necessary to give special thought to the choice of evaluation method. First, due to the relatively small number of examples, we used leave-one-out cross-validation instead of k-Fold cross-validation. Though computationally more expensive, leave-one-out is a better estimator for small datasets \cite{wong2015performance}. 

Secondly, the class imbalance meant that optimizing for classification accuracy could result in sub-optimal (or even degenerate) models, because it implicitly assigns higher weights to examples of the majority class. To address this, the true positive rate (sensitivity) and true negative rate (specificity) were tracked separately. As a single evaluation metric for choosing one model over another, we use the {\em balanced misclassification loss}  (see Appendix \ref{append_bal_loss}).

%k-Fold cross-validation is only a good estimator when $\frac{n}{k}$ is fairly larger than $k$ \cite{wong2015performance} where n is the number of examples
%Every example in the dataset was left out once, and the model fit to the remaining data.
%For example, in our problem, using 10-fold cross validation would mean only about 18 examples in each fold, leading to high variance in estimates of the error. We thus applied Leave-one-out cross validation which is more suited for small datasets \cite{wong2015performance}. Every example in the dataset was left out once, and the model fit to the remaining data.

\section{Results}
\label{section_results}
%The objective of our experiments was to build predictive models of the breathing patterns extracted during a 5 minute period of spontaneous breathing trial (SBT) prior to extubation of a preterm newborn. Data from a total of 186 patients (136 successes and 50 failures) were available. We show results of modeling experiments as well as of classification.
Of the 186 patients recruited, 136 succeeded extubation and 50 failed. We discuss the results of our experiments in the following sub-sections.

\begin{table}[t]
\centering 
\caption{The type and parameters of the distributions of best fit to the dwell (or sojourn) times in each respiratory pattern for success and failure patients.}
\label{table_dwell_dist}
\begin{tabular}{lcc}
\toprule
 & \textbf{Success} & \textbf{Failure} \\
\hline
\textbf{Pause} & Exponential  &  Exponential   \\
 & $\mu=2.51$ &  $\mu = 2.94$   \\
\cmidrule{2-3}
\textbf{Asynchrony} & GeneralizedExtremeValue & GeneralizedExtremeValue \\ 
& k=0.63, $\sigma$ = 1.30, $\mu$=1.85  & k=0.65, $\sigma$ = 1.36, $\mu$=1.81  \\
\cmidrule{2-3}
\textbf{Movement} & GeneralizedPareto &   GeneralizedPareto      \\
& k=-0.22, $\sigma$ = 3.62  & k=-0.11, $\sigma$ = 3.31  \\
\cmidrule{2-3}
\textbf{Synchrony} & InverseGaussian &   InverseGaussian \\
& $\mu$ =8.61, $\lambda$ = 3.61 & $\mu$ =7.83, $\lambda$ = 3.41 \\
\cmidrule{2-3}
\textbf{Unknown} & GeneralizedPareto &   GeneralizedPareto      \\
& k=-0.07, $\sigma$ = 2.07  & k=-0.10, $\sigma$ = 2.05  \\
\bottomrule
\end{tabular}
\end{table}

\begin{table}[b]
\caption{Respiratory state transition probabilities for the Success population modeled as a Semi-Markov chain. }
\label{table_state_transition_success}
\centering %\resizebox{\columnwidth}{!}
{
\begin{tabular}{cccccc}%|c|c|c|c|c|c|}
\toprule
 & \textbf{PAU} & \textbf{ASB} & \textbf{MVT} & \textbf{SYB} & \textbf{UNK}\\
\hline
\textbf{PAU} & 0  &  0.27   & 0.09  &  0.26   & {\color{red} \textbf{0.38}}\\
%\hline
\textbf{ASB} & 0.10  &       0  & 0.16  &   0.29  &   \textbf{0.45} \\
%\hline
\textbf{MVT} &0.12  &   0.32       &  0  & {\color{red} \textbf{0.43}}  &   0.14  \\
%\hline
\textbf{SYB} & 0.06  &   0.25  &   0.15     &    0  &  \textbf{0.54} \\
%\hline
\textbf{UNK} & 0.13  &0.28 &   0.04    & \textbf{0.55}      &    0 \\
\bottomrule
\end{tabular}
}
\end{table}

%(Bold, black font indicate the most probable transitions that were same in both populations. Bold, red indicate most probable transitions that were different)
\begin{table}[b]
\caption{Respiratory State Transition probabilities for Failure population modeled as semi-Markov chain. }
\label{table_state_transition_failure}
\centering %\resizebox{\columnwidth}{!}
{
\begin{tabular}{cccccc}%|c|c|c|c|c|c|}
\toprule
 & \textbf{PAU} & \textbf{ASB} & \textbf{MVT} & \textbf{SYB} & \textbf{UNK}\\
\hline
\textbf{PAU} & 0  &  0.28   & 0.06  &  {\color{red} \textbf{0.39}}   & 0.28\\
%\hline
\textbf{ASB} & 0.12   &       0  &  0.21  &   0.28  &   \textbf{0.40} \\
%\hline
\textbf{MVT} &0.17  &   {\color{red} \textbf{0.41}}        &  0   & 0.32  &   0.09  \\
%\hline
\textbf{SYB} & 0.14  &   0.21  &   0.14    &    0  &  \textbf{0.52} \\
%\hline
\textbf{UNK} & 0.15  &0.30 &   0.03    & \textbf{0.52}      &    0\\
\bottomrule
\end{tabular}
}
\end{table}

\subsection{Analysis of Respiratory State Durations}
As a starting point, the total time spent in each breathing pattern was obtained for the two groups of patients, shown in Fig.~\ref{fig_state_duration}. It was normalized by the total duration of spontaneous breathing to account for the difference in number of examples in both classes. In the 2 groups, SYB was the dominant state at least 50\% of the time, with the success patients spending slightly more time in this state. The failure patients on the other hand spent a greater fraction of time in the PAU state. The time spent in the other states were essentially equivalent as confirmed by standard error estimates obtained by bootstrapping, shown in Fig. \ref{fig_state_duration}.

\subsection{Modeling of Dwell Times}

The optimal dwell time distributions in each state were estimated for both groups (as described in section \ref{section_semi_markov_chain}) and are summarized in Table \ref{table_dwell_dist}. The detailed plots of the probability density functions (PDF) are shown in Appendix \ref{append_dwell_dists}. It was interesting to observe that for each pattern, the {\em distribution type} which best fit the dwell times in both populations were same. We suspect that this an indication of some underlying consistency in breathing behaviour of premature infants in spite of extubation outcome. It should also be noted that the dwell time was distributed exponentially only in the PAU state. As discussed earlier, the use of a Markov chain model would have implicitly taken an exponential distribution for all states. The use of semi-Markov framework has thus allowed for a more expressive and accurate representation.

\subsection{Modeling Transitions}
\label{subsec_modeling_transitions}
The respiratory pattern sequences were modeled as semi-Markov chains (Tables \ref{table_state_transition_success} and \ref{table_state_transition_failure}). Each cell in the matrix represents the probability of transitioning from the state labeled on the row to that on the column. It was observed that the most probable transition given ASB and SYB (the breathing states) was the same in both infants who succeeded and those who failed extubation (shown in bold, black font). However, it differed for PAU and MVT states (shown in bold, red font). This suggests that in terms of transition behaviour, the 2 groups of infants differ more in the transitions emanating from non-breathing states than from breathing states.

Further, the symmetric KL divergence ($D_{KLS}$) between the 2 transition matrices for the semi-Markov model was estimated as 0.27. When modeled as a Markov chain (see details in Appendix \ref{append_markov_transition_mat}), the probabilities of cross-pattern transitions and the $D_{KLS}$ tended towards zero (0.0019), indicating that the learned models were almost identical due to relatively long dwell times. The semi-Markov transition matrix, on the other hand, being invariant to sampling frequency changes and/or length of dwell time, resulted to better numerical resolution of cross-pattern transitions and the learning of more discriminating characteristics between both groups of infants.
%an increase from the Markov chain case suggesting that more discriminating information was learned.

%The data sequences were modeled first as a Markov chain. Due to the relatively long dwell times, the probabilities of cross-pattern transitions (off-diagonal elements) were all near-zero (details in Appendix \ref{append_markov_transition_mat}). The symmetric KL divergence between the  success and failure transition distributions was 0.0019, indicating the distributions were almost identical.

\subsection{Generative Classification with Semi-Markov Chain Models}
The classification performance of the learned semi-Markov transition models was evaluated using leave-one-out cross validation. As described in section \ref{section_semi_markov_chain}, the likelihoods of each test example were computed using its entire sequence of transitions (Lk-ALL). Failure patients were identified at a rate (sensitivity) of 50\% while specificity was 73\%. %These models take a generative approach to classification. Considering that we have almost 3 times the more success than failure patients, we suspect that the low sensitivity might be due to the dominance of the distribution learned for the success patients.

The predictive value of the individual patterns/states was further evaluated. In particular, to compute the likelihood of a test sequence based on one state, the product of cross-state transitions emanating from only that state are taken. As before, this likelihood is computed with respect to the transition models for the 2 classes, and a prediction is made by selecting the class whose model gave higher likelihood. Results are shown accordingly in Table \ref{table_result_summary} where {\em Lk-STATE} represents prediction made using likelihood of the "STATE" specified. The highest sensitivity of 68\% was obtained by the PAU, MVT, SYB states. Overall, the PAU pattern gave the lowest loss of 0.37.

\subsection{Discriminative Classification with SVM}

The following features, motivated by the semi-Markov chain model, were extracted from each subject and applied in an SVM classifier.

\begin{itemize}
\item Total dwell time in each respiratory pattern as a fraction of the total sequence duration, ({\em Dw-All}) - 5 features
\item Number of transitions from pattern $i$ to pattern $j$ (where $i\neq j$) as a fraction of the total dwell time in pattern $i$, $\forall i \in S$, ({\em Tr-All}) - 20 features
\item Number of occurrences of each respiratory pattern as a fraction of the number of occurrences of all patterns, ({\em Oc-All}) - 5 features
\end{itemize}

\setlength{\textfloatsep}{10.5pt plus 1.0pt minus 2.0pt}

\begin{table}[!b]
\caption{Performance of classification methods. Lk-ALL or Lk-STATE refers to classification using likelihood of chain considering all states or a specified "STATE". Dw, Oc, Tr refer to features extracted based on dwell time, occurrence count and transitions in states.}
\label{table_result_summary}
\centering
\begin{tabular}{lccc}
    \toprule
    %\multicolumn{1}{c}{} & \multicolumn{2}{c}{Performance}\\
    %\cmidrule{2-3}
    Approach & Sensitivity & Specificity & Loss\\
    \midrule
    \textbf{Generative (Semi-Markov)}\\
    \midrule
    %Lk-ALL (Markov) & 0.46 & 0.73 & 0.40\\
    Lk-ALL & 0.50 & 0.73 & 0.38\\
    Lk-PAU & \textbf{0.68} & \textbf{0.58} & \textbf{0.37}\\
    Lk-ASB & 0.63 & 0.48 & 0.45\\
    Lk-MVT & 0.68 & 0.50 & 0.41\\
    Lk-SYB & 0.68 & 0.53 & 0.40\\
    Lk-UNK & 0.61 & 0.44 & 0.48\\
    \midrule
    \textbf{Discriminative (SVM)}\\
    \midrule
    Dw-ALL & 0.18 & 0.93 & 0.45 \\
    Oc-ALL & 0.44 & 0.69 & 0.44 \\
    Tr-ALL & 0.82 & 0.41 & 0.39 \\
    Dw-Oc-Tr-ALL & 0.64 & 0.63 & 0.37\\
    Dw-Oc-Tr-PAU  & \textbf{0.84} & \textbf{0.54} & \textbf{0.31}\\
    Dw-Oc-Tr-ASB  & 0.38 & 0.75 & 0.44 \\
    Dw-Oc-Tr-MVT  & 0.60 & 0.58  & 0.41 \\
    Dw-Oc-Tr-SYB   & 0.26 & 0.81 & 0.46\\
    Dw-Oc-Tr-UNK  &  0.62 & 0.43 & 0.48 \\
    %\midrule
%     \textbf{Reference (SVM)}\\
%     \midrule
%     Cardiorespiratory metrics  & 0.83 & 0.74 & 0.22\\
   \bottomrule
\end{tabular}
\end{table}

The features were first used individually (Dw-ALL, Oc-ALL, Tr-ALL) and then as an ensemble (Dw-Oc-Tr-ALL) of 30 features. Similar to the generative case, the predictive value of each individual pattern/state was evaluated - for each state, the dwell time in that state, {\em Dw-STATE} (1 feature), the cross-transitions, {\em Tr-STATE} (4 features) and the occurrence count, {\em Oc-STATE} (1 feature) were combined ({\em Dw-Oc-Tr-STATE}) to train the classifier. We applied SVMs with RBF/Gaussian kernels.
%, which have 2 hyper-parameters: the box constraint $C$ which controls the intensity of regularization and kernel scale $\gamma$ which controls the width of the Gaussian. 
Leave-one-out cross-validation was used in a grid search to find the best pair of hyper-parameter (box constraint $C$ and kernel scale $\gamma$) values based on the balanced misclassification loss. This grid search was repeated for each feature set since each would have different optimal values of $C$ and $\gamma$.

\begin{figure}[t]
  \centering 
  \includegraphics[width=0.48\textwidth]{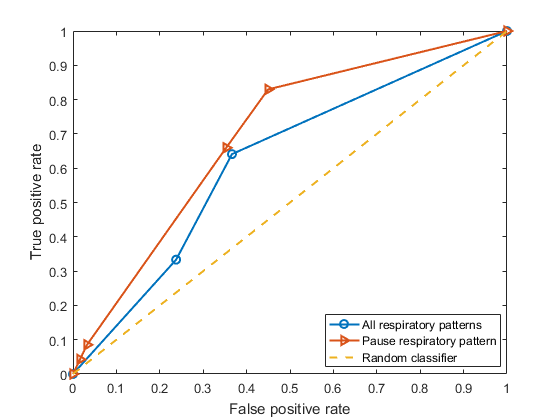} 
  \caption{Receiver-operating characteristic (ROC) curve for support vector machine trained on summary features of all respiratory patterns (AUC=0.62) and on summary features of only the Pause (PAU) pattern (AUC=0.70)}
  \label{fig_svm_roc} 
\end{figure}

When using features for all patterns, the sensitivity and specificity at the optimal hyper-parameter setting were 64\% and 63\%, respectively. The highest sensitivity of 84\% was obtained when using features of only the Pause pattern, which also gave the lowest loss of 0.31. It could also be observed that whereas PAU and MVT patterns gave higher sensitivities, SYB and ASB gave higher specificities. This is likely an indication that the Pause and Movement patterns characterise better patients who may fail extubation while the breathing patterns better characterise patients who succeed. All results are summarised in Table \ref{table_result_summary}. In Fig. \ref{fig_svm_roc}, we show the receiver-operating characteristic (ROC) curve for the 2 SVM models which gave the lowest loss. ROC was obtained by fixing $C$ at the optimal value and varying $\gamma$.

\setlength{\textfloatsep}{10.5pt plus 1.0pt minus 0.0pt}
%\pagebreak

\section{Discussion} 
\label{section_discussion}

We demonstrated the practical application of semi-Markov chains for modeling and classification of respiratory pattern behaviour of preterm infants in the immediate period prior to extubation. We showed that semi-Markov chain models provide more expressive and robust details about the underlying time series compared to Markov chain models. In terms of dwell time behaviour, the model revealed consistency between the success and failure groups in all respiratory states. Differences were highlighted primarily in transition behaviour arising from the Pause and Movement Artifact patterns. Prediction results confirmed that these 2 patterns provide more discriminating information (especially for patients who failed extubation) than any other pattern.

Indeed, using only features from the Pause pattern led to the greatest detection of failed extubations (with a sensitivity of 84\%), but failed to recognizing nearly half of the successful extubations (specificity of 54\%). This means that at the time infants were deemed ready for extubation, 8 out of 10 failures would be prevented, but at the expense of unnecessarily prolonging the course of mechanical ventilation in 5 out of 10 successes. Previous work \cite{precup2012} which applied SVM (on a much smaller cohort) using cardiorespiratory variables gave comparable sensitivity but a higher specificity of 74\%. It will be necessary to repeat those experiments with this larger dataset. Another viable path for future work is in training a mixture of experts model (MEM) that classifies failure patients based on Pause and Movement states, and success patients on the breathing states. The use of automatically extracted respiratory patterns for prediction of extubation readiness provides an approach that unveils intuition and enhances interpretable models. We emphasize that all babies used in this study were deemed ready for extubation by an attending clinician, so these results constitute an improvement in detecting problem cases over current practice. The advantage of using an automated approach is that we can provide a quantified, repeatable and precise analysis breathing patterns to support clinical decisions.

%\pagebreak

\bibliographystyle{IEEEtran}
\bibliography{main}

\pagebreak

\appendix
%\section*{Appendices}
\subsection{Balanced Misclassification Loss}
\label{append_bal_loss}

Accuracy is defined as:
\begin{eqnarray}
acc &=& \frac{tp + tn}{tp + fn + tn + fp}\\
&=& \frac{tp + tn}{p + n} = \frac{tp}{p + n} + \frac{tn}{p + n}\\
& =& \frac{tp}{p} \left (\frac{p}{p + n} \right ) + \frac{tn}{n} \left (\frac{n}{p + n} \right)\\
\label{eqn_acc_weighted_sum}
&=& sens.\left (\frac{p}{p + n} \right ) + spec. \left (\frac{n}{p + n} \right)
\end{eqnarray}
where $tp$, $tn$, $fp$ and $fn$ are respectively the number of true positives, true negatives,  false positives and false negatives; $p$ is the number of positive examples and $n$ the number of negative examples. $sens$ is sensitivity and $spec$ is specificity.

This means that, if the class proportions are 75:25, a classifier which simply predicts the majority class would have a misleading accuracy of 75\%. The {\em balanced accuracy} measure, $acc_b$,  corresponds to the average of the sensitivity and specificity measures:
\begin{equation}
\label{eqn_acc_eq_weighted_sum}
acc_b = sensitivity * 0.5 + specificity * 0.5
\end{equation}

%\pagebreak

\subsection{Transition Matrices for Fitted Markov chain Model }
\label{append_markov_transition_mat}

The Markov chain transition matrices for the success and failure populations are shown in Tables \ref{table_state_transition_success_m} and \ref{table_state_transition_failure_m}, respectively. It can be seen that due to extremely long dwell times in states compared to cross-state transitions, the diagonal elements (self-transitions) account for nearly all of the probability on each row.

\begin{table}[h]
\caption{Respiratory state transition probabilities for the Success population modeled as a Markov chain}
\label{table_state_transition_success_m}
\centering %\resizebox{\columnwidth}{!}
{
\begin{tabular}{cccccc}%|c|c|c|c|c|c|}
\toprule
 & \textbf{PAU} & \textbf{ASB} & \textbf{MVT} & \textbf{SYB} & \textbf{UNK}\\
\hline
\textbf{PAU} & \textbf{0.9936}  &   0.0019  &   0.0004  &   0.0022 &    0.0018\\
%\hline
\textbf{ASB} & 0.0006  &   \textbf{0.9953} &    0.0010  &   0.0013  &   0.0018 \\
%\hline
\textbf{MVT} &0.0012  &   0.0029  &   \textbf{0.9931} &    0.0020  &   0.0007 \\
%\hline
\textbf{SYB} & 0.0003  &   0.0005  &   0.0003  &  \textbf{ 0.9977} &    0.0012 \\
%\hline
\textbf{UNK} & 0.0015  &   0.0031 &    0.0003  &   0.0055  &   \textbf{0.9895} \\
\bottomrule
\end{tabular}
}
\end{table}

\begin{table}[h]
\caption{Respiratory State Transition probabilities for Failure population modeled as Markov chain}
\label{table_state_transition_failure_m}
\centering %\resizebox{\columnwidth}{!}
{
\begin{tabular}{cccccc}%|c|c|c|c|c|c|}
\toprule
 & \textbf{PAU} & \textbf{ASB} & \textbf{MVT} & \textbf{SYB} & \textbf{UNK}\\
\hline
\textbf{PAU} & \textbf{0.9920}  &   0.0022 &    0.0007  &   0.0020  &   0.0032\\
%\hline
\textbf{ASB} & 0.0005  &   \textbf{0.9955}  &   0.0007  &   0.0013  &   0.0020 \\
%\hline
\textbf{MVT} &0.0007  &   0.0021   &  \textbf{0.9934}  &   0.0028 &    0.0010 \\
%\hline
\textbf{SYB} & 0.0001  &   0.0005  &   0.0003  &   \textbf{0.9978}   &  0.0012 \\
%\hline
\textbf{UNK} & 0.0013 &    0.0027 &    0.0004  &   0.0056  &   \textbf{0.9899} \\
\bottomrule
\end{tabular}
}
\end{table}

\pagebreak
\subsection{Dwell/Sojourn Time Distribution Fits }
\label{append_dwell_dists}
Plots of the probability density functions (PDF) of sojourn times in all states are shown in Fig 4, as well as the distributions of best fit based on the bayesian information criterion (BIC).
\begin{figure}[h]
  \centering 
  \includegraphics[width=.48\textwidth]{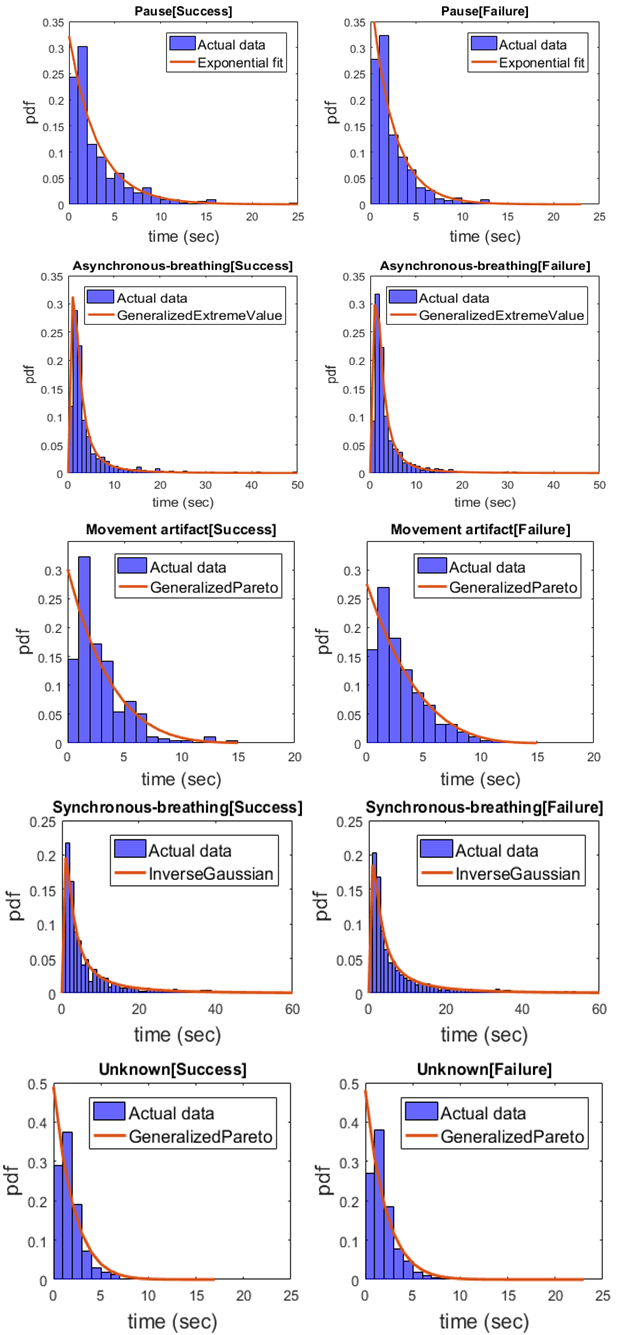} 
  %\caption{Probability Density Functions (PDF) of Dwell Time Distributions in all 5 respiratory patterns for success and failure patients}
  \label{fig_dwell_dists} 
\end{figure}
\\
Fig 4: Probability Density Functions (PDF) of Dwell Time Distributions in all 5 respiratory patterns for success and failure patients

% \begin{figure*}[t]
%   \centering 
%   \includegraphics[width=1\textwidth]{figs/transition_prob_pau_mvt.png} 
%   \caption{Given a start state of Pause (left) and Movement Artifact(right), the figures show the probabilities of transitioning to other states/patterns.}
%   \label{fig_transition_prob_pau_mvt} 
% \end{figure*}
%Appendices
%1. all sojourn distributions
%2. transition matrix for markov chain
%3. graphical view of transition matrix for all start states inn semi-markov chain.

\end{document}